\documentclass{article}

\usepackage{arxiv}

\usepackage[utf8]{inputenc}
\usepackage[T1]{fontenc} 
\usepackage{hyperref}       
\usepackage{url}            
\usepackage{booktabs}      
\usepackage{amsfonts}       
\usepackage{nicefrac}       
\usepackage{microtype}      
\usepackage{graphicx}
\usepackage{natbib}
\usepackage{doi}
\usepackage{hyperref}
\usepackage{wrapfig}
\usepackage{multirow}
\usepackage{geometry}
\usepackage{amsmath}
\DeclareUnicodeCharacter{2212}{-}

\title{Bag of Lies: Robustness in Continuous Pre-training BERT}
\date{} 			

\author{ \href{https://orcid.org/0000-0003-1565-4077}{\includegraphics[scale=0.06]{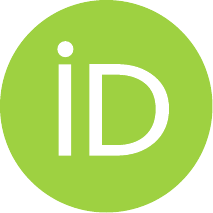}\hspace{1mm}Ine Gevers} \\
	CLiPS\\
	University of Antwerp\\
	Antwerp, 2000 \\
	\texttt{ine.gevers@uantwerpen.be} \\
	\And
	\href{https://orcid.org/0000-0002-9832-7890}{\includegraphics[scale=0.06]{orcid.pdf}\hspace{1mm}Walter Daelemans} \\
	CLiPS\\
	University of Antwerp\\
	Antwerp, 2000 \\
	\texttt{walter.daelemans@uantwerpen.be} \\
}

\renewcommand{\shorttitle}{\textit{arXiv} Template}

\hypersetup{
pdftitle={Bag of Lies: Robustness in Continuous Pre-training BERT},
pdfsubject={q-bio.NC, q-bio.QM},
pdfauthor={Ine Gevers, Walter Daelemans},
pdfkeywords={Continued pre-training, adversarial attacks, robustness, COVID-19},
}

\begin{document}
\maketitle

\begin{abstract}
This study aims to acquire more insights into the continuous pre-training phase of BERT regarding entity knowledge, using the COVID-19 pandemic as a case study. Since the pandemic emerged after the last update of BERT's pre-training data, the model has little to no entity knowledge about COVID-19. Using continuous pre-training, we control what entity knowledge is available to the model. We compare the baseline BERT model with the further pre-trained variants on the fact-checking benchmark Check-COVID. 
To test the robustness of continuous pre-training, we experiment with several adversarial methods to manipulate the input data, such as training on misinformation and shuffling the word order until the input becomes nonsensical.
Surprisingly, our findings reveal that these methods do not degrade, and sometimes even improve, the model's downstream performance. This suggests that continuous pre-training of BERT is robust against misinformation. 
Furthermore, we are releasing a new dataset, consisting of original texts from academic publications in the LitCovid repository and their AI-generated false counterparts.

\end{abstract}

\section{Introduction}
While pre-trained Large Language Models achieve remarkable results on a variety of downstream tasks, it is also known that their performance decreases when they are applied to tasks relying on information outside of the scope of their original pre-training distribution \citep{oren-etal-2019-distributionally}. Standard practice to alleviate this issue is to continue pre-training the models (i.e., Masked Language Modeling on large unlabeled text data sets) before fine-tuning (i.e., using smaller labeled task-specific data after pre-training). This technique aims at bridging the gap between a model's original knowledge and the specialized information required for its current application. For instance, continuous pre-training (CPT) has shown its merits for specialised in-domain applications (e.g., \citet{lee2020biobert,chalkidis-etal-2020-legal}).\\

We aim to explore the level and robustness of entity knowledge that BERT can acquire through CPT on topics diverging from its original pre-training data. Focusing on a case study enables us to isolate and examine the specific impact of CPT on entity knowledge, sidestepping the confounding factors that often complicate such analyses.\\ 
We focus on entity knowledge regarding the COVID-19 pandemic, a topic that emerged after the end of BERT's initial pre-training phase. Although BERT's original dataset may include abstract knowledge about viruses and diseases, the specifics of the COVID-19 pandemic present a novel challenge. By leveraging a COVID-19 fact-checking benchmark, Check-COVID \citep{wang-etal-2023-check-covid}, as our evaluative framework, we aim to shed light on questions surrounding the stability and robustness of knowledge acquisition during CPT.\\
Our methodology examines various factors that could influence the efficacy of CPT, including the size of the data set (cf. \citet{rietzler2019adapt}), the veracity of information, the source of the data, the degree to which the training data is aligned with the task data (cf. \citet{gururangan-etal-2020-dont}), the word order within the data, and model size (regarding data memorization, cf. \citet{kharitonov2021bpe}). The original data sources we examine are academic publications (from the LitCovid repository \citep{litcovid}), task-adaptive data (from the fact-checking benchmark Check-COVID \citep{wang-etal-2023-check-covid}), and social media data (from Reddit). 

\begin{wrapfigure}[15]{r}{0.4\textwidth}
    \vspace{-20pt}
    \centering
    \includegraphics[width=0.4\textwidth]{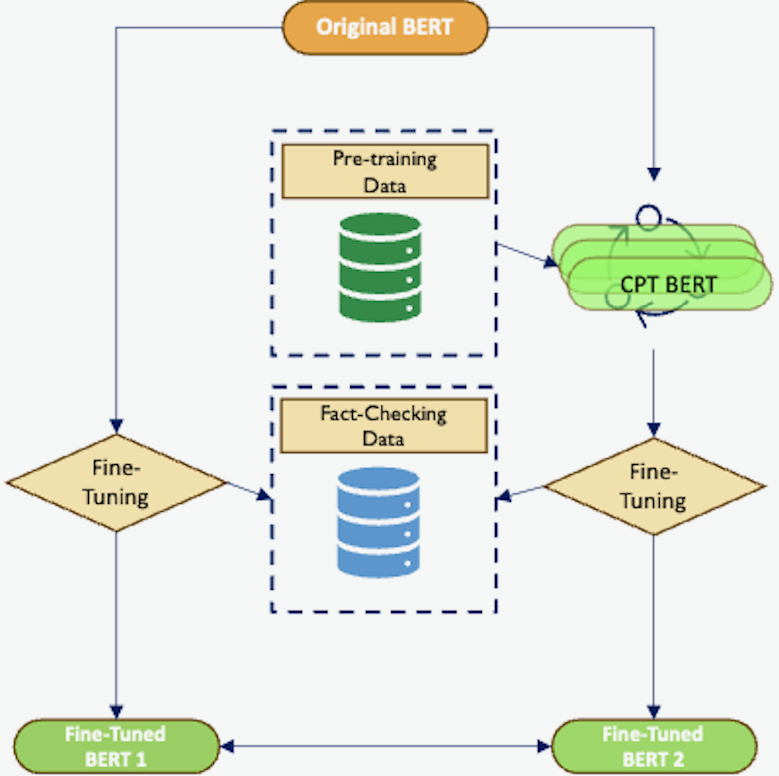}
    \caption{Illustration of the model architecture.}
    \label{fig:model_architecture}
\end{wrapfigure}

We employ two adversarial techniques to manipulate the original input data: (1) misinformation; and (2) shuffling the word order. 
We CPT BERT on the diverse forms of input data related to COVID-19, consequently fine-tuning and evaluating the models' performance on the Check-COVID benchmark. Figure \ref{fig:model_architecture} illustrates the experimental setup described in this study.
Among the key findings of our study are the positive effects of CPT on downstream performance, and the surprising robustness of the models against adversarial techniques, with certain adversarial inputs even enhancing the model's performance. 
We plan to release the dataset we created for this purpose, which contains the texts extracted from academic publications in LitCovid paired with their AI-generated misinformation and AI-generated paraphrasing, and to make our code publicly available.\\

\section{Related Work}
Knowing what information is used by models at inference time is crucial to ascertain the model's trustworthiness and generalization. However, uncovering this is a major challenge: a whole field of study, BERTology, is dedicated entirely to studying the inner workings of transformers such as BERT~\citep{rogers2021primer}. While this field is concerned with a variety of knowledge, such as syntactic, semantic, or world knowledge, others are mainly interested in uncovering what factual information is used by BERT.\\ 
For instance, \citet{podkorytov2021can} analyse the internal components of BERT that are responsible for the output, to measure factual knowledge present in the transformer model and its ability for generalization in downstream tasks. They find that BERT's knowledge is fragile, and based on token co-occurrence in the pre-training data. Conversely, \citet{petroni2019language} convert factual triplets (subject-relation-object) into prompts to probe factual knowledge and find that BERT has a strong ability to recall factual information. However, \citet{https://doi.org/10.1002/widm.1518} find that this approach is not fool-proof, since adding negation to the prompt distorts the results. Using Masked Language Modeling (MLM) probing, \citet{penha2020does} also observe that BERT stores knowledge about books, movies, and music in its parameters.\\
As can be seen, there is some discrepancy in the reported results about how and what factual knowledge is stored in BERT. This could be an effect of the variety of evaluation metrics used to measure knowledge in BERT. 
Previous research has used both intrinsic and extrinsic evaluations. Intrinsic variants include fill-in the gap probes in Masked Language Modeling (MLM); using self-attention weights; and probing classifiers using different BERT representations as input~\citep{rogers2021primer}. Extrinsic evaluations on downstream NLP tasks have been carried out on benchmarks such as CREAK \citep{onoe2021creak} which tests for entity knowledge. We propose fact-checking tasks as an additional extrinsic evaluation.

While results do not always agree about the extent to which knowledge is reliably incorporated in the parameters of BERT, it is generally agreed that the textual data used to (pre-)train the transformer model has a significant role in the acquisition of that knowledge. Using large language models' perplexity on masked spans in texts about entities that are excluded from the original pre-training data, \citet{onoe2022entity} demonstrate that models struggle with making inferences about unseen entities, from which can be derived that the knowledge about these entities is limited.\\ 
Updating the available knowledge, then, is fundamental to improving the models' performance. However, pre-training the model from scratch is computationally expensive and time-consuming \citep{lamproudis2021developing}, but relying on extensive fine-tuning can lead to catastrophic forgetting~\citep{recadam}. Therefore, intermediate techniques such as continuous pre-training (CPT) could alleviate this issue~\citep{cossu4495233continual}.\\ 
CPT is especially important for events that took place after the last update of the model, or for specific topics that are not well represented in the original training data (i.e., in-domain: biomedical BioBERT \citep{lee2020biobert}, legal: LEGAL-BERT \citep{chalkidis-etal-2020-legal}). \citet{lemmens-etal-2022-contact} continue pre-training the Dutch RobBERT model on COVID-19 related Tweets, and show that the resulting model outperforms the original model on vaccine hesitancy detection.
Additionally, \citet{gururangan-etal-2020-dont} show that task-adaptive pre-training, in which the downstream task's unlabeled data is used for CPT, is a promising method compared to using large amounts of in-domain data. 
However, the amount of data needed for CPT is domain-dependent~\citep{rietzler2019adapt}. Most research uses human-generated data as input to pre-train transformer models, but given the increasing rise of generative language models such as the GPT-family, some research has experimented with using AI-generated data, showcasing its effectiveness~\citep{eldan2023tinystories}.\\
Although CPT is now standard practice, questions persist about its effectiveness and optimal configurations.~\citep{bacco2023instability}. Also, the stability of the process has been questioned since even one sentence can alter the model's downstream performance \citep{bacco2023instability}, and fine-tuning the model on a large dataset can obliterate the effects of CPT~\citep{zhu2021does}.\\ 
As such, efforts have been made to test the boundaries by exploring adversarial techniques. Literature shows that using nonsensical input texts (i.e., randomly selected n-grams, non-human language, or different word order) in CPT does not lead to worse results~\citep{chiang2020pre,krishna-etal-2021-pretraining-summarization,sinha-etal-2021-masked}. It is hypothesized that pre-training mainly teaches the model hierarchical structures, long-distance dependencies, and higher-order word co-occurrences, for which the distributional information of the input text is enough. Some research notes that adding noise to the input data, which is argued to encourage the diversity of the embedding vectors, even helps downstream performance~\citep{wang2019improving}. However, to the best of our knowledge, no one experimented with using factually incorrect data as a confounding factor, which is especially relevant for fact-checking tasks: models' pre-training data could include unverified information, so if misinformation influences the models' output, this needs to be addressed.

\section{Research questions and hypotheses}\label{research questions and hypotheses}
This study is guided by the following research questions and hypotheses:
\begin{enumerate}
    \item \textbf{Does BERT utilize entity knowledge for fact verification?} We investigate whether BERT is capable of leveraging specific entity knowledge to verify facts within given statements during fine-tuning. To measure this, we focus on entity knowledge that was not present in the original pre-training data, but introduced during the CPT phase. 
    We hypothesize that BERT can benefit from new entity knowledge seen during CPT to perform fact verification tasks. Additionally, task-adaptive pre-training will enhance performance: using CPT data that is more aligned with the specific language use of the downstream task is beneficial~\citep{gururangan-etal-2020-dont} 
    \item \textbf{Is the veracity of that entity knowledge important for the accuracy of fact verification by BERT?} This question aims to understand the impact of the truthfulness of the provided entity knowledge on the model's performance.
    We expect the presence of erroneous information to negatively affect the model's ability to accurately verify facts during fine-tuning. In like manner, using questionable sources as input data such as Reddit will also decrease performance.
    \item \textbf{How robust is the CPT phase?} We examine whether CPT still helps performance on fact verification when the input data is manipulated to confuse the model (i.e., misinformation, shuffled word order).
    We hypothesize that the CPT phase is not robust when it comes to misinformation, but in accordance with prior work, we assume that the use of nonsensical data (i.e., shuffled word order) should not decrease the results on downstream tasks~\citep{sinha-etal-2021-masked,krishna-etal-2021-pretraining-summarization,chiang2020pre}. Following previous literature, we assume that small amounts of CPT data will already show differences in downstream tasks (see \citet{bacco2023instability}), but we hypothesize that larger pre-training datasets will make these effects more robust, with more correct information leading to better performance and more incorrect information resulting in worse outcomes.
\end{enumerate}

It is important to note that the primary goal of this study is not to surpass the current state-of-the-art (SOTA) models in fact-checking: we do not expect that by CPT alone we could match the current SOTA. Rather, we focus on the effect of adding new entity knowledge in the CPT phase, and a fact-checking setup gives us a controlled environment to evaluate the importance of this entity knowledge in a downstream task that revolves around entity knowledge. However, insights gained from this research could assist other techniques focused on improving fact-checking performances. This includes understanding the potential benefits of using limited pre-training data, evaluating the significance of the data source for pre-training (e.g., the use of AI-generated data and data from Reddit), the impact of using misinformation during CPT, the possible gains from task-adaptive pre-training, and evaluating the process' robustness by manipulating the word order of the input data. 

\section{Methods}
\subsection{Data}\label{Data}
\begin{figure}[h] 
    \begin{center} 
    \includegraphics[width=1.0\textwidth]{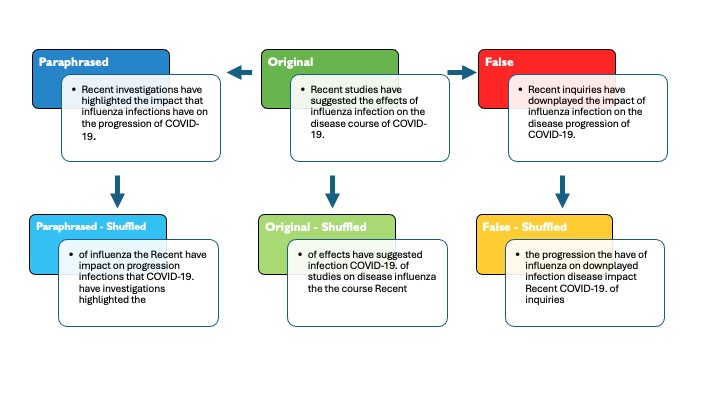} 
    \vspace{-10pt}
    \caption{Illustration of adversarial transformations of the input text.}
    \label{adversarial}
    \end{center}
    \vspace{-10pt}
\end{figure}

We use various sources as the starting point for creating the input data to continue pre-training BERT. In this section, we describe these sources, and the subsequent transformations we applied to test the robustness of the CPT process. Figure \ref{adversarial} demonstrates the implemented procedures through examples. Additionally, we describe the benchmark we use to fine-tune and evaluate the resulting CPT models on. 

As a starting point, we extract texts from the LitCovid repository, which contains academic publications about the COVID-19 pandemic~\citep{litcovid}. These texts, following the structure of the dataset, can be titles, abstracts, or entire paragraphs from original academic publications. 
To verify the impact of the CPT data size, we compare a model with CPT on 200 text units with one on 10,000 texts\footnote{We experimented with larger sizes up to 1 million texts from the original LitCovid data, but because 1) the performance plateaued early and 2) we could not generate this many texts using GPT-4, we restrict our discussion to the 200 and 10,000 variants.}.\\
Following \citet{gururangan-etal-2020-dont}, we also implement task-adaptive pre-training. For this purpose, we use the unlabeled text data from the fact-checking benchmark as input during CPT, prior to fine-tuning the model on the labeled data. To verify that topical vocabulary plays a role, we add a baseline model with CPT on similar task data, but other topics. To this end, we gather data from the fact-checking benchmarks Liar \citep{wang-2017-liar} and VitaminC \citep{schuster-etal-2021-get}, and we filter out all data points that mention 'covid', 'pandemic', 'corona', 'covid-19', or 'coronavirus'.\\
The last original data source we consider in this study is user-generated data on the social media platform Reddit\footnote{\url{https://www.kaggle.com/datasets/pavellexyr/the-reddit-covid-dataset?select=the-reddit-covid-dataset-comments.csv}}. The dataset contains posts and comments mentioning COVID-19, from which we randomly sample 10,000 comments to be used as input data.\\
There are two adversarial techniques that we apply to modify the original input data. For the academic texts (derived from LitCovid), we use the GPT-4Turbo API to revert the truthfulness of the text, artificially generating misinformation. To ensure that the text's veracity is important besides the specific AI-generated language use, we also use GPT-4Turbo to paraphrase the original data. 
For all text sources (i.e., LitCovid, Reddit, task-adaptive, and AI-generated), we shuffle the word order to distort the text. The shuffling is done both inside one line of text as well as for the entire collection of texts.\\
We evaluated and compared the readability\footnote{Flesch Reading Ease} and complexity of the resulting datasets (i.e., LitCovid, paraphrased, and misinformation for both 200 and 10,000 samples, and shuffled). We note that larger datasets exhibit lower lexical diversity (TTR) and higher complexity (measured by Shannon entropy and perplexity). 
Shuffling word order has minimal impact on text complexity and diversity. Comparing human to AI-generated texts, we observe that perplexity decreases and readability increases from original to misinformation to paraphrased text, while entropy remains consistent.

We evaluate the resulting CPT models on the same benchmark, Check-COVID~\citep{wang-etal-2023-check-covid}. The benchmark combines claims from newspaper articles\footnote{either "extracted": copied verbatim from the newspaper article, or "composed": rewritten claim by an annotator based on the content of the newspaper article}, annotated by experts, and evidence from the CORD-19 repository \citep{wang-etal-2020-cord} to test LLM's fact-verification abilities concerning the COVID-19 pandemic. In total, there are 1,500 claims (evenly distributed over the labels 'support', 'refute', and 'not-enough-information'). For more statistics about the benchmark, please see \citep{wang-etal-2023-check-covid}.
While the approach presented in their research focuses on rationale selection from texts for the task, our own study, in contrast, aims to examine the role of entity knowledge in BERT's CPT data, for which the fact-checking setup offers a good case study. For our purposes, we concentrate on the 'support' and 'refute' labels and exclude the 'not-enough-information' label. To overcome variability issues observed when evaluating on the original test set only, we apply 5-fold cross-validation. As mentioned earlier, we do not compare our results to those reported on the fact-verification task; we do not expect our results to be competitive with this task. Rather, we focus on the relative performance difference between a base model without entity knowledge, and models that have access to entity knowledge through CPT.

\subsection{Experiments}
Similar to previous work (e.g., \citet{bacco2023instability}), we focus on the BERT-base model~\citep{devlin-etal-2019-bert}. We deliberately choose a smaller, older model, because it has been used in previous research concerning CPT, making it easier to situate and compare our work. Additionally, the smaller size allows us to run our experiments locally in a manageable time frame.\\
We run three baseline models to measure the influence and robustness of entity knowledge added during CPT. Specifically, we use the BERT-base model\footnote{\url{https://huggingface.co/google-bert/bert-base-uncased}}, an in-domain pre-trained model BioBERT \citep{lee2020biobert}, and a BERT-base model CPT on task-adaptive but non-COVID topics (following \citet{gururangan-etal-2020-dont}). 
We use a fact-checking setup because it is a good case to measure the model's ability to learn from data about an entity inserted in CPT. The binary classification ("support" and "refute") offers a straightforward way to evaluate models' performance. By keeping the rest of the setup identical, and by only changing the input data for CPT, we can pinpoint the relative impact of this entity knowledge on the model's performance. 

\subsubsection{Continue pre-training BERT}
As mentioned before, we do our main experiments with BERT-base, but given that model size is an important variable for data memorization (cf. \citet{kharitonov2021bpe}), we add additional results with BERT-large\footnote{\url{https://huggingface.co/google-bert/bert-large-uncased}}\footnote{We cannot compare with BioBERT, since this model only has BERT base as underlying model.}. For model specifications, we refer to Appendix \ref{Appendix D 2} 
After further pre-training, we fine-tune and evaluate the resulting models on the down-stream task. We report results across 5 random seeds in 5-fold cross-validation, indicating the average performance as well as the standard deviation across the seeds.

\subsubsection{Prompt GPT API to revert truthfulness}
In order to create the adversarial data, in which we artificially generate misinformation, we leverage the OpenAI GPT-4 API. We use the original texts extracted from the LitCovid repository as input. 
We experimented with several variations of the prompt. For transparency, we add the final version of the prompt in Appendix \ref{Appendix A}. 
Since simple negations such as "not" are known to confuse BERT-base models~\citep{truong-etal-2023-language}, we take care to instruct the generative model to go beyond simple negations: we include this specifically in the prompt, and we found that by increasing the temperature setting, the model complies better with this demand. We use the gpt-4-turbo-preview\footnote{\url{https://platform.openai.com/docs/models/gpt-4-and-gpt-4-turbo}, model date: 25/01/2024} model, setting the model temperature to 0.5. We keep the output length approximately as long as the original input text\footnote{We use the tiktoken package (\url{https://pypi.org/project/tiktoken/0.1.1/}) to estimate the number of input tokens.} and we manually verify the quality of a sample of the generated output. 
We use the same model and technique to generate paraphrased data from the original texts, the prompt can be found in Appendix \ref{Appendix B}.  
We release the resulting paired dataset\footnote{The cost of generating the misinformation dataset was \$148.30, the paraphrased dataset \$111.96.} (i.e., the original texts sampled from the academic publications in LitCovid \citep{litcovid}, and the generated counterparts)\footnote{We will release the dataset.}. 
We focus on a case study about COVID-19, an entity that is not present in the original pre-training data for various reasons. Mainly, investigating an entity that is present in the original data would require us to delete parts of BERT's original pre-training data, which could introduce unexpected variance into the model's performance. Alternatively, pre-training a model from scratch excluding the relevant entity knowledge is both time-intensive and resource-heavy, making it an impractical approach for our research objectives. 

\section{Results and discussion}

\begin{table*}[h!]
  \centering
  \setlength{\tabcolsep}{4pt} 
  \renewcommand{\arraystretch}{1.1} 
  \begin{tabular}{|l||c|c|c||c|c|c|}
    \hline
    \multirow{2}{*}{\textbf{Model}} & \multicolumn{3}{c||}{\textbf{BERT-base}} & \multicolumn{3}{c|}{\textbf{BERT-large}} \\ \cline{2-7}
    & \textbf{Macro F1} & \textbf{Precision} & \textbf{Recall} & \textbf{Macro F1} & \textbf{Precision} & \textbf{Recall}\\ \hline
    BERT-baseline                           & 61.76 (±3.52)          & 63.80       & 62.56 & 65.56 (±2.88) & 68.24 & 66.40          \\ \hline
    BioBERT                                & 66.48 (±3.43)*          & 67.36          & 66.72   & / & / & /        \\ \hline
    Task-adaptive other topic              & 62.64 (±2.16)           & 64.64          & 63.60  & 63.16 (±2.95) & 65.84 & 64.32          \\ \hline \hline
    Task-adaptive Check-COVID              & 64.96 (±2.73)*          & 66.96          & 65.76   & 65.04 (±3.45) & 66.48 & 65.88        \\ \hline
    Task-adaptive Check-COVID shuffled     & 62.60 (±4.67)           & 65.60          & 64.12   & 65.00 (±1.61) & 66.44 & 65.64        \\ \hline
    LitCovid200 (True)                     & 58.76 (±3.48)           & 62.04          & 60.12   & 65.12 (±2.49) & 66.72 & 65.96       \\ \hline
    LitCovid200 (True) shuffled            & 64.60 (±3.03)*          & 66.56          & 65.20   & 64.88 (±2.99) & 66.52 & 65.56         \\ \hline
    LitCovid200 (Paraphrased)              & 64.92 (±2.99)*          & 67.04          & 65.64   & 64.88 (±2.36) & 66.56 & 65.52        \\ \hline
    LitCovid200 (Paraphrased) shuffled     & 64.52 (±2.58)*          & 66.48          & 65.20   & 65.08 (±2.74) & 66.80 & 65.76       \\ \hline
    LitCovid200 (False)                    & 64.80 (±2.76)*          & 66.40          & 65.32   & 63.56 (±2.57) & 65.56 & 64.56        \\ \hline
    LitCovid200 (False) shuffled           & 64.16 (±2.61)*          & 66.00          & 65.08   & 64.48 (±2.57) & 67.00 & 65.40       \\ \hline
    LitCovid10K (True)                     & 60.72 (±2.40)           & 62.28          & 61.40   & 66.28 (±1.42) & 67.76 & 66.76        \\ \hline
    LitCovid10K (True) shuffled            & 63.24 (±5.99)*          & 65.12          & 64.28   & 64.08 (±7.27) & 65.72 & 65.44        \\ \hline
    LitCovid10K (Paraphrased)              & 64.32 (±2.99)*          & 66.56          & 65.36   & 65.64 (±1.86) & 67.00 & 66.08       \\ \hline
    LitCovid10K (Paraphrased) shuffled     & 65.08 (±1.12)*          & 67.12          & 65.76   & 66.56 (±2.92) & 68.60 & 67.28        \\ \hline
    LitCovid10K (False)                    & 63.88 (±3.04)*          & 65.64          & 64.76   & 66.00 (±1.41) & 67.36 & 66.44        \\ \hline
    LitCovid10K (False) shuffled           & 64.00 (±0.98)*          & 66.40          & 65.12   & 64.28 (±6.41) & 66.88 & 65.44       \\ \hline
    Reddit                                 & 64.56 (±2.71)*          & 67.16          & 65.52   & 65.48 (±1.91) & 66.36 & 65.92        \\ \hline
    Reddit shuffled                        & 64.60 (±4.57)*          & 66.92          & 65.56   & 65.72 (±4.02) & 67.88 & 66.68        \\ \hline
    \end{tabular}
    \caption{Model performance on Check-COVID. We compare three baseline models (BERT, BioBERT, task-adaptive model on other topics) with the CPT models. We report the average result of the models in 5-fold cross-validation across 5 random seeds. We indicate the relative standard deviation for the Macro F1-score across the seeds. '*' denotes whether the difference in macro F1 performance from the baseline BERT model is statistically significant. Following prior work, we use the McNemar test for this purpose. If $\alpha$ \textless 0.05, we can assume that the model is significantly different from the baseline. The effect sizes (calculated with Cohen's g) are small to medium. For the exact measures, we refer to Appendix \ref{Appendix E}.}
    \label{tab:model_performance}
\end{table*}

We report the performance of the models on Check-COVID in Table \ref{tab:model_performance}. We focus first on BERT-base, since this is also the model discussed in prior literature. Our analysis shows that CPT generally helps performance on this downstream task, which confirms prior literature. 
Specifically, CPT on in-domain (i.e., biomedical) data~\citep{lee2020biobert} and task-adaptive pre-training~\citep{gururangan-etal-2020-dont} significantly improve downstream performance. However, improvements with task-adaptive pre-training are contingent on the data being topically aligned. Simply having similar data from tasks, but about different topics, is insufficient to gain significant improvements.\\ 
Surprisingly, using accurate information extracted from academic publications does not improve performance, and this holds true even when larger datasets are employed. However, the BERT-base models CPT on the generated misinformation (both on the smaller and larger datasets) significantly improved compared to the baseline model and the models CPT on the original academic texts. Also the models CPT on AI-generated paraphrased data improve. So, using AI-generated text helps in this context, but the veracity of the text plays no role: there is no significant difference between BERT-base models CPT on correct or incorrect AI-generated text.\\
We hypothesize that this can be explained by the language use of AI-generated texts being more "standard", which helps BERT to learn the relevant patterns~\citep{eldan2023tinystories}. Also, the data analysis in Section \ref{Data} showed that AI-generated texts have a lower perplexity compared to the original human generated text, which could be a reason why the BERT-base model performs better with this data. An alternative explanation could be that there is not yet enough data provided during CPT, and that with more data a breaking point will be observed. We leave this for future research to explore.\\ 
To verify that not only AI-generated language is responsible for this remarkable result, we use Reddit comments about COVID-19 -of which the veracity of the content can be questioned- as input data. We observe that also in this setup, the resulting model significantly outperforms the baseline model. This unexpected finding could lead to further research in domains with restricted data availability: user-generated data from social media platforms are generally omitted in these contexts exactly because of their questionable veracity and quality, but this result could indicate that including social media data is a viable option.\\
Consistent with earlier studies, the shuffling of word order does not significantly affect downstream performance in most scenarios. However, we observe a notable exception: in instances where CPT does not yield improvements over the baseline model (i.e., when correct information from academic texts is used), the shuffling of this data leads to improved performance. Additionally, when the two adversarial attacks used in this study (i.e., misinformation and shuffling word order) are combined, we observe that the resulting models still outperform the baseline. This could suggest that CPT is rather robust, and primarily focuses on learning associations on document level. If the language or writing style is too noisy or deviates significantly from the language used in the task (as described in \citet{gururangan-etal-2020-dont}), making it more challenging for the model to learn (as is the case with academic publications), then reformatting it using generative AI techniques and/or shuffling the word order can potentially aid the model to generalize.\\
We repeated our experiments using BERT-large to measure the effect of model size. First, there is no significant difference in performance between the BERT-large based CPT models: CPT does not bring improvements, but adversarial attacks also do not degrade model performance. This is unexpected, since larger models are generally associated with more data memorization \cite{kharitonov2021bpe}, which could have lead to models CPT on incorrect information performing worse. Second, there are no significant differences when comparing the paired BERT-base and BERT-large models (e.g., BERT-base CPT on Reddit and BERT-large CPT on Reddit). Thus, while model size mitigates the effects of CPT, the model is still robust against adversarial attacks.

In summary, we can answer the research questions from Section \ref{research questions and hypotheses} as follows:
\begin{enumerate}
    \item \textbf{Does BERT utilize entity knowledge for fact verification?} Relevant entity knowledge generally helps BERT's downstream performance. However, we do observe that the language use of the input data should be aligned to the task data: as expected, task-adaptive pre-training yields improvements, but contrary to our initial expectations, using academic texts from LitCovid do not improve results. However, a larger model does not significantly benefit from CPT.
    \item \textbf{Is the veracity of that entity knowledge important for the accuracy of fact verification by BERT?} No, using misinformation or questionable data sources as input does not degrade the model's performance. On the contrary: there is a significant improvement compared to the baseline performance for BERT-base.
    \item \textbf{How robust is CPT in enhancing the model's ability for fact verification?} We find that CPT is robust against the two adversarial techniques we present in this work (i.e., misinformation and shuffling word order), also when combined. Using larger data sizes shows the same conclusions as smaller data sizes.
\end{enumerate}

\section{Conclusion}
Continuous pre-training has become a standard practice for addressing the limitations of language models for niche or not well-represented areas, or to update a model's information after the initial pre-training. Nevertheless, the stability of this process has been questioned, highlighting the need for further investigation into its reliability and impact on model performance~\citep{bacco2023instability}. 
In this study, we examine a specific aspect of CPT by focusing on entity knowledge. While considerable research efforts have investigated in-domain pre-training (e.g., \citet{lee2020biobert,chalkidis-etal-2020-legal}), few have looked at entity knowledge: to the best of our knowledge, the benchmark CREAK is one of its kind investigating common sense reasoning over entity knowledge~\citep{onoe2021creak}. However, we propose using fact-checking benchmarks as a means to assess a model's grasp on entity knowledge.\\
In this case study, we focus on the COVID-19 pandemic. Since the pandemic emerged after the last update of BERT's pre-training data, the model has little to no entity knowledge about COVID-19. Using CPT, we control what entity knowledge is available to the model. We compare the baseline BERT model with the CPT variants on the fact-checking benchmark Check-COVID~\citep{wang-etal-2023-check-covid}.
We compare three baseline models (i.e., a vanilla BERT \citep{devlin-etal-2019-bert}, an in-domain pre-trained BioBERT \citep{lee2020biobert}, and a task-adaptive model on other topics (based on \citep{gururangan-etal-2020-dont}) with BERT models CPT on relevant entity knowledge. For this, we use three data sources: academic publications (the LitCovid repository \citep{litcovid}), task data (the unlabeled texts from CheckCovid \citep{wang-etal-2023-check-covid}), and social media (Reddit). Further, we compare performance of two model sizes: BERT-base and BERT-large.\\ 
Since the robustness of the CPT process is sometimes questioned, we explore two adversarial attacks that manipulate this input data: deliberately using misinformation (which we apply to LitCovid, generating misinformation with the OpenAI GPT-4 API. We publicly release the resulting paired dataset), and shuffling the word order (which we apply to all three data sources).\\ 
Consequently, we compare the baseline models with the CPT models by fine-tuning and evaluating them on the same fact-checking benchmark Check-COVID. We apply McNemar tests on the models' predictions to confirm significance and cohen's g to report effect size. A manual error analysis on a sample of the models' output did not reveal any distinct patterns.\\
Surprisingly, our findings indicate that the veracity of the text is not an important factor. BERT-base models CPT on AI-generated data perform better than BERT-base models CPT on original correct information, but there is no significant difference whether that AI-generated data is correct (paraphrased) or incorrect (misinformation). Additionally, using a source of questionable content quality (i.e., Reddit) also improves BERT-base performance. Consistent with prior results, shuffling word order has no effect~\citep{chiang2020pre,krishna-etal-2021-pretraining-summarization,sinha-etal-2021-masked}. However, we note that in the cases where CPT does not lead to improvements on the baseline performance (i.e., when correct information from academic texts is used), shuffling the word order of that data results in significantly better performance. We observe that even when the two adversarial attacks are combined, this does not have a negative effect on the downstream performance. We observe that a larger model size (i.e., BERT-large) is less impacted by CPT, but is still robust against the adversarial attacks.\\
Looking ahead, we suggest several avenues for further research, including an examination of the internal representations within the models similar to the methods proposed by \citet{bacco2023instability} and the utilization of larger CPT datasets. In our approach, we leverage the GPT-4 API to generate paraphrases and misinformation, but there is a potential variability and superficiality in the outputs. Future research might benefit from creating texts manually to compare with the automated outputs of GPT-4. Further, this case study demonstrates that user-generated data from social media platforms, despite their questionable veracity and quality, can be used as input data for CPT. This could inspire future work on domains with restricted data availability to use social media data. We also like to point out that this research is conducted on a small controlled case-study, looking at the COVID-19 pandemic, and that generalizations to other topics or domains should be investigated in further research. \\

\section{Limitations}
While our work is, to the best of our knowledge, the first to tackle misinformation in CPT, there are certain limitations that were not addressed in this case-study. 
First, since it is a case-study, this exploratory study is limited to one entity (COVID-19), and one downstream dataset (Check-COVID). This is partly due to a lack of qualitative labeled COVID-19 fact-checking data. After considering the possibilities, we decided to do an in-depth study on one topic, and include other topics in future research. However, it is thus possible that results could vary for other topics.
Second, the data size used for CPT is relatively limited (up to 10,000 texts): using more data during CPT could affect the trends observed in this study. However, in our experiments we noted no difference between models CPT on 10,000 or 1 million texts from the original LitCOVID repository\footnote{We did not go beyond 10,000 texts for the AI-generated counterparts because of resource limitations.}. Additionally, we compared model performances using incrementally more texts during CPT (each time, we added 500 texts), but the performance of the models plateaued quickly.
Third, the relationship between CPT and entity knowledge is indirect, and the fact-checking task is used as an intuitive proxy for this question. It could for instance be possible that the CPT models did not retain information from the CPT dataset. More internal analyses, such as probing, would give more insights. 

\section*{Acknowledgments}
This research was made possible with a grant from the Fonds Wetenschappelijk Onderzoek (FWO) project 42/FA030100/9770.

\bibliographystyle{unsrtnat}
\bibliography{references}

\appendix
\section{Appendix}\label{Appendix A}
Prompt to generate misinformation from original texts in academic publications present in the LitCovid repository:\\
\\
You are an AI language model tasked with processing academic texts related to COVID-19. Your objective is to rewrite an Input Text A but sustain the opposite position.\\
In simpler terms, let's imagine there are two statements (A and B) related to COVID-19, and they cannot both be true at the same time. If the Input Text supports fact A, the task is to rewrite the text in a way that now supports statement B.\\
Let’s first understand the problem by reading the instructions, then extract relevant variables, and make a plan. Then, let’s carry out the plan, calculate intermediate variables (pay attention to commonsense), solve the problem step by step, and show the answer. Look at the example cases below to understand the task. Each example consists of an example input (Input Text), what your output should look like (Expected Output), and an example of undesirable output (Wrong Output). After having read and understood the examples, transform the Input Text.

**Instructions:**

1. **Read the input text carefully.** It will be an abstract or a paragraph from an academic publication about COVID-19.

2. **Transform the text.** Your goal is to reverse the truthfulness of the information presented in the text. If the Input Text supports a statement, make sure the Output Text supports the opposite. 

3. **Maintain academic tone and style.** Despite the transformation, the resulting text should preserve the formal and structured nature of academic writing.

4. **Make sure the output is coherent.** While making an Output Text contradictory to the Input Text, make sure the Output Text is coherent and logical. 

5. **Use various linguistic techniques.** Avoid relying solely on negations. Employ a range of linguistic strategies such as rephrasing with antonyms, altering contexts, introducing contrary facts, or any creative method that inverts the factual basis of the content. 

6. **Output format:** Return the transformed text as a string. If multiple texts are provided in one session, separate each transformed text with a newline. 

7. **Length of response:** The Output Text should be approximately the same length as the Input Text to ensure that the essence and detail of the original content are mirrored in the transformation.

**Example Case 1:**\\
- **Input Text:** "Recent studies indicate that COVID-19 primarily spreads through respiratory droplets."
- **Expected Output:** "In-depth analyses suggest that COVID-19's transmission is unrelated to respiratory droplets."
- **Wrong Output:** "Old studies indicate that COVID-19 does not spread through respiratory droplets."\\\
**Example Case 2:**\\
- **Input Text:** "In a groundbreaking discovery, researchers have identified a specific protein that plays a crucial role in the severity of COVID-19 symptoms. Understanding the interaction of this protein with the virus could lead to targeted therapeutic interventions and improved outcomes for patients."
- **Expected Output:** "In a groundbreaking discovery, researchers have refuted the existence of a specific protein that plays a crucial role in the severity of COVID-19 symptoms. Since no specific protein interacts with the virus dismisses, there is no possibility of targeted therapeutic interventions, challenging the potential for improved outcomes for patients."
- **Wrong Output:** "In an unoriginal discovery, researchers have not identified a specific protein that plays an uncrucial role in the severity of COVID-19 symptoms. Disregarding the interaction of this protein with the virus could not lead to targeted therapeutic interventions and improved outcomes for patients."\\\

Read the instructions and example cases carefully. Only once you fully comprehend your task, proceed with transforming the provided COVID-19 academic text(s) according to these instructions.

\section{Appendix}\label{Appendix B}
Prompt to generate paraphrases from original texts in academic publications present in the LitCovid repository:\\

**Context:** You are an AI language model tasked with processing academic texts related to COVID-19. Your role involves creatively paraphrasing these texts. The original texts may include abstracts or paragraphs from academic publications. Your objective is to paraphrase its content using a variety of linguistic techniques. This exercise aims to explore the flexibility of language and understand how the same information can be presented in various ways while maintaining logical coherence and readability.

**Instructions:**

1. **Read the input text carefully.** It will be an abstract or a paragraph from an academic publication about COVID-19.

2. **Transform the text.** Your goal is to paraphrase the information presented in the text. Avoid using only synonyms. Employ a range of linguistic strategies such as changing word classes, using a different grammatical structure or voice (active vs. passive), elaborating on the original text, or any creative method that paraphrases the content. 

3. **Maintain academic tone and style.** Despite the transformation, the resulting text should preserve the formal and structured nature of academic writing.

4. **Make sure the output is coherent.** While paraphrasing, make sure the output is coherent and logical. 

5. **Do not only use synonyms.** Rely on other paraphrasing techniques besides using synonyms of terms used in the original sentence.

6. **Output format:** Return the transformed text as a string. If multiple texts are provided in one session, separate each transformed text with a newline. 

7. **Length of response:** The transformed text should be approximately the same length as the input text to ensure that the essence and detail of the original content are paraphrased in the transformation.

**Example:**

- **Input Text:** "Recent studies indicate that COVID-19 primarily spreads through respiratory droplets."
- **Expected Output:** "Recent research shows that COVID-19 mainly transmits via respiratory droplets."

Read the context and instructions carefully. Only once you fully comprehend your task, proceed with transforming the provided COVID-19 academic text(s) according to these instructions.

\section{Appendix}\label{Appendix D}

\subsection{Model specifications continual pre-training BERT} \label{Appendix D 2}
To continue pre-training BERT, we follow this procedure.
We pre-train for one step using the MLM objective, for which we use the baseline script on HuggingFace (which was also used for the research in \citep{gururangan-etal-2020-dont}. As is standard practice, we mask 15\% of the tokens. The learning rate was set at 5e-05, consistent with the usual rate for domain adaptation~\citep{bacco2023instability,gururangan-etal-2020-dont}. We train for one epoch, using mixed precision (fp16) to accelerate the process. Since we use a collection of text units as input data, we opt for line-by-line, which directs the model to use the text inputs as separate sequences. When the CPT is completed, we save and upload the model to HuggingFace, making it accessible for further use.\\

\subsection{Model specifications fine-tuning BERT on fact verification} \label{Appendix D 3}
To fine-tune BERT on the downstream task of fact verification, we follow this procedure.
We maintain stable hyperparameter settings to ensure consistency in the experimental conditions (similar to \citet{bacco2023instability}). Hyperparameter tuning was conducted on the baseline BERT model using the Optuna library~\citep{optuna_2019}. We set the learning rate to 3.5e−05, adjust the batch size to 32 to accommodate hardware limitations, use 5 training epochs, and implement early stopping after two epochs to prevent overfitting.

\section{Appendix} \label{Appendix E}
In Table \ref{tab: effect sizes}, we give the p-values and effect sizes when comparing the BERT-base baseline to BERT-base CPT models. 

\begin{table}[h!]
    \centering
    \begin{tabular}{l|c|c}
    models & p-value &  Cohen's g \\
    \hline
    bert - biobert                          & 0.010 &                         0.20 \\
    bert - litcov200 True shuffled          & 0.020 &                         0.20 \\
    bert - litcov200 False                  & 0.001 &                         0.18 \\
    bert - litcov200 False shuffled         & 0.010 &                         0.21 \\
    bert - litcov200 paraphrased            & 0.010 &                         0.17 \\
    bert - litcov200 paraphrased shuffled   & 0.007 &                         0.19 \\
    bert - checkcovid                       & 0.007 &                         0.17 \\
    bert - reddit                           & 0.010 &                         0.20 \\
    bert - reddit shuffled                  & 0.040 &                         0.17 \\
    bert - litcov10K False                  & 0.030 &                         0.14 \\
    bert - litcov10K paraphrased            & 0.030 &                         0.14 \\
    bert - litcov10K paraphrased shuffled   & 0.002 &                         0.18 \\
    \hline
    \end{tabular}
    \caption{P-values (McNemar) and effect sizes (Cohen's g) comparing BERT-base baseline and BERT-base CPT models.}
    \label{tab: effect sizes}
\end{table}

\end{document}